\documentclass{article}

\PassOptionsToPackage{numbers,compress,square,sort,comma}{natbib}

\usepackage[preprint]{neurips_2019}

\usepackage[utf8]{inputenc} 
\usepackage[T1]{fontenc}    
\usepackage{hyperref}       
\usepackage{url}            
\usepackage{booktabs}       
\usepackage{amsfonts}       
\usepackage{nicefrac}       
\usepackage{microtype}      
\usepackage{graphicx}
\graphicspath{ {./imgs/} }
\usepackage[toc,page]{appendix}
\usepackage{caption}

\title{Searching for an \textit{(un)stable equilibrium}: experiments in training generative models without data}

\author{%
  Terence Broad \\
  Department of Computing\\
  Goldsmiths, University of London\\
  \texttt{t.broad@gold.ac.uk} \\
   \And
   Mick Grierson \\
   Creative Computing Institute \\
   University of the Arts London \\
   \texttt{m.grierson@arts.ac.uk} \\
}

\begin{document}

\maketitle

\begin{abstract}

This paper details a developing artistic practice around an ongoing series of works called \textit{(un)stable equilibrium}. These works are the product of using modern machine toolkits to train generative models without data, an approach akin to traditional generative art where dynamical systems are explored intuitively for their latent generative possibilities. We discuss some of the guiding principles that have been learnt in the process of experimentation, present details of the implementation of the first series of works and discuss possibilities for future experimentation. 

\end{abstract}

\section{Introduction}

In this work we are utilising toolkits for data driven optimisation (such as PyTorch \cite{paszke2017pytorch}) and pre-existing generative model architectures with strong inductive biases, to explore the latent possibilities of potential generative outputs that don't mimic any existing data distribution. We have set out to achieve this by finding ways of `training' these generative models without any training data. This approach can be seen as akin to practices in traditional generative art, where dynamic systems are built and the role of the artist is to design or influence this process to some degree, based on intuition and exploration \cite{mccormack2004generative}. We see this as a continuation of artistic practice describe by Bense as \textit{Generative Aesthetics}, where we are using the modern tools of gpu-optimised linear algebra libraries, differentiable objective functions and gradient-based optimisation to design and explore the characteristics of these new `aesthetic structures' \cite{bense1965projekte}.

\section{Guiding Principles}

Here we discuss some of the useful concepts and fruitful techniques that have been learnt through the process of experimentation and utilised in the works in \textit{Series 1} (see Section \ref{case-study} for details). 

\subsubsection*{Complexity | Stochasticity}

A lot of the trial and error in the practice is finding the right balance of complexity and stochasticity. Often finding the right batch size is key, too low and gradients quickly explode, too high and the error signal averages out any potential system dynamics, resulting in stasis.

\subsubsection*{Relational Constraints}

We often utilise constraints that are relative to the output of a given batch. These constraints may be distances in embedding spaces using techniques from metric learning \cite{kulis2013metric}, or measure of diversity in pixel space of a generators batch-wise output.

\subsubsection*{Exploiting Boundaries}

We exploit the small differences in the way different differentiable functions measure distance and difference.  These discrepancies can be exploited to create internal system dynamics that continually inject a level of randomness into the training dynamics.

\subsubsection*{Diametric Optimisation}

In some of the arrangements for the works in \textit{Series 1}, we train some of the networks in the system with two diametrically opposed loss functions (propagated after exposure to different batches to prevent them from completely cancelling out). While this may be counter-intuitive from the perspective of optimisation, it provides an anchor of stability in networks ensembles where the other networks are relying on that network's output in the process of training.

\subsubsection*{Discovering (Un)stable Equilibria}

The previous guiding principles are all techniques that serve the goal of finding a balance of randomness and stability: to find an equilibrium in the space of potential system dynamics which is stable enough to prevent gradients collapsing or exploding, but unstable enough to produce unexpected results.

\section{Case Study:  \textit{Series 1}}
\label{case-study}

For the works in \textit{Series 1}, the setup resembles the popular generative adversarial networks \cite{goodfellow2014generative} ensemble, however here we have two generators (both using progressively-growing, style-based generator architectures \cite{karras2019style}). The `discriminator' sometimes acts in the traditional way as a binary classifier, trying to correctly classify which generator has produced which image. Alternatively, it is sometimes trained simultaneously with both diametrical opposing adversarial loss functions (this is true for the the works \textit{1:1}, \textit{1:2} and \textit{1:6}). In either case, the discriminator's classification output, and distance measurements in the discriminator's embedding space are utilised for training the generators. 

The generators compete in having their output as being recognized as the output of the other network, either using the classification output of the discriminator or by having the distance of their embeddings (in discriminator space) as close as possible to the other generators. Both generators also compete to have more variety in the colours they output at pixel level (in their respective batch) than the other generator. These arrangements result in abstract, sometimes orthogonal compositions from the two generators. After training is completed, the resulting images from the two generators is presented side-by-side, as a video piece showing a synchronised interpolation between their respective latent spaces (see Figures \ref{fig1}-\ref{fig6} in the Appendix for stills from and links to the video works in the series).

\section{Discussion and Future Work}
\label{discussion}

In this work, we have developed a practice that relies heavily on the subjective aesthetic analysis of the output. Through subjective interpretation of the output of these systems (often through closely monitoring results visually throughout training), an intuition has been developed that has informed decisions in the iterations of model design. We find this practice to resonate strongly with Stanley's description of the role of artistic understanding in the process of researching artificial systems \cite{stanley2018art}. 

In future experiments, we look to develop the practice further and explore aesthetic possibilities of other commonly used techniques in machine learning, such as the variety of diversity metrics now used to assess generative models, such as the inception score \cite{salimans2016improved} and Fréchet inception distance \cite{heusel2017gans}. We also want to experiment with integrating meta-information of training performance into the training of the model, as well as adaptive and evolutionary techniques to dynamically change the model architectures and meta-model arrangements.

\subsubsection*{Acknowledgments}

This work has been supported by UK’s EPSRC Centre for Doctoral Training in Intelligent Games
and Game Intelligence (IGGI; grant EP/L015846/1).

\bibliographystyle{unsrt}
\bibliography{neurips_2019}

\newpage
\appendixpage

\begin{figure}[!htb]
  \centering
\includegraphics[width=0.9\textwidth]{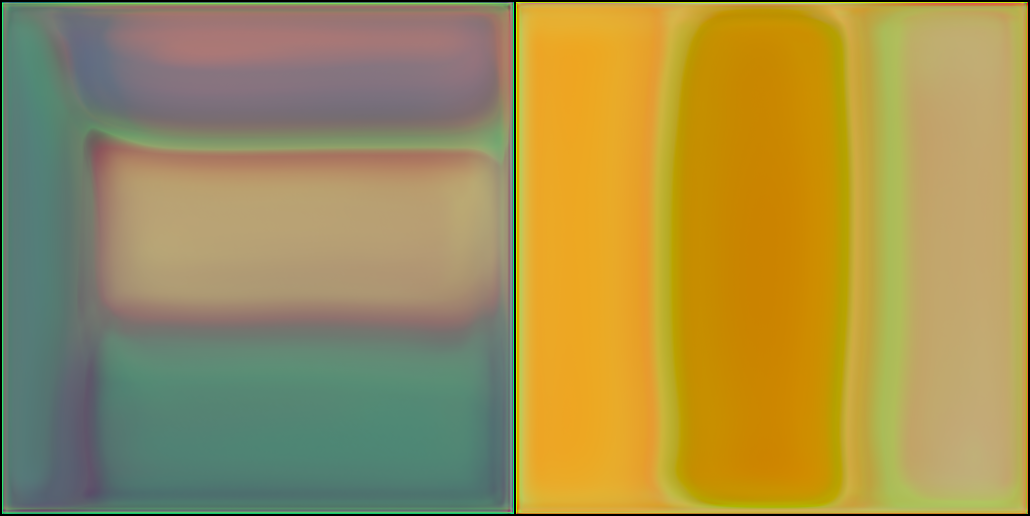}
  \caption{Still from \textit{(un)stable equilibrium 1:1}. The work can be viewed at: \url{https://youtu.be/r6MB555mXXM}}
  \label{fig1}
\end{figure}

\begin{figure}[!htb]
  \centering
\includegraphics[width=0.9\textwidth]{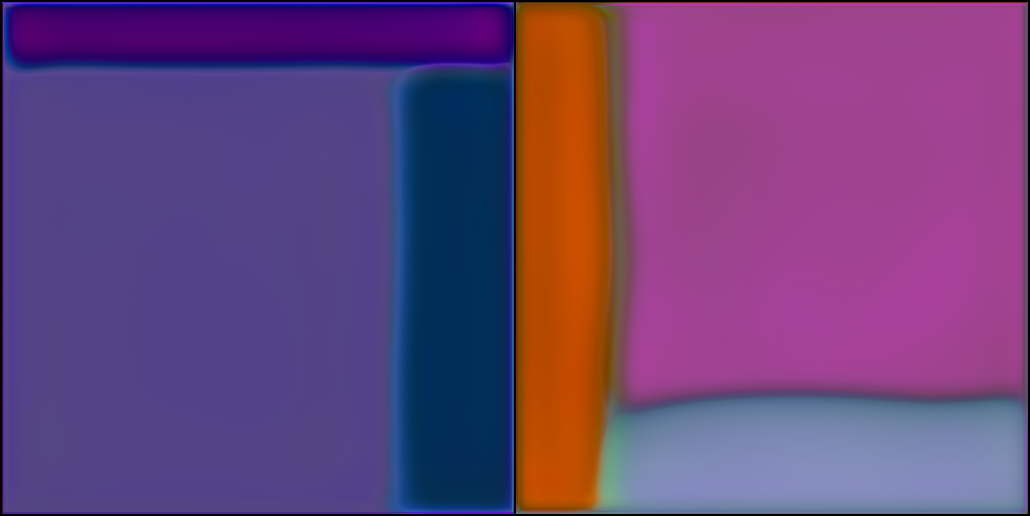}
  \caption{Still from \textit{(un)stable equilibrium 1:2}. The work can be viewed at: \url{https://youtu.be/P_LLD8ffgVc}}
  \label{fig2}
\end{figure}

\newpage

\begin{figure}[!htb]
  \centering
\includegraphics[width=0.9\textwidth]{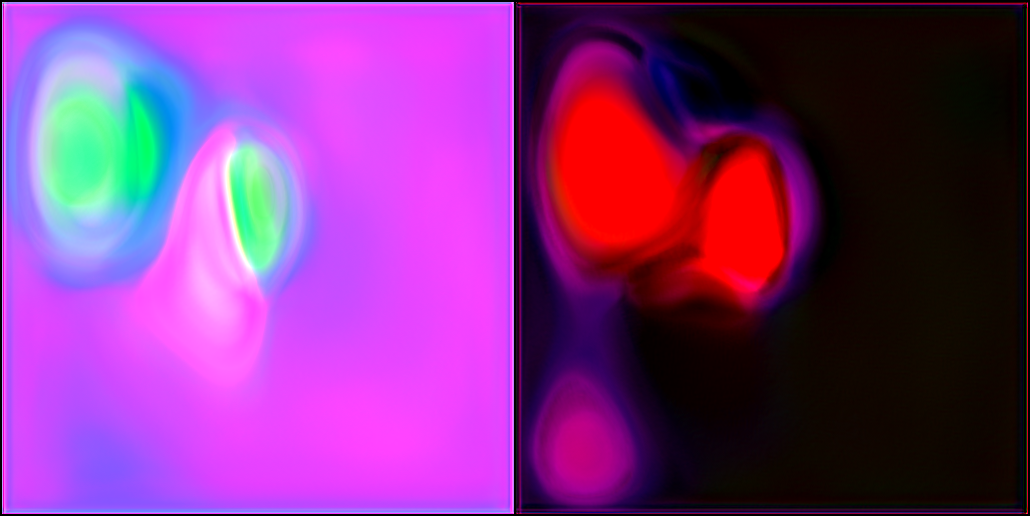}
  \caption{Still from \textit{(un)stable equilibrium 1:3}. The work can be viewed at: \url{https://youtu.be/A4CsL5TAvQU}}
  \label{fig3}
\end{figure}

\begin{figure}[!htb]
  \centering
\includegraphics[width=0.9\textwidth]{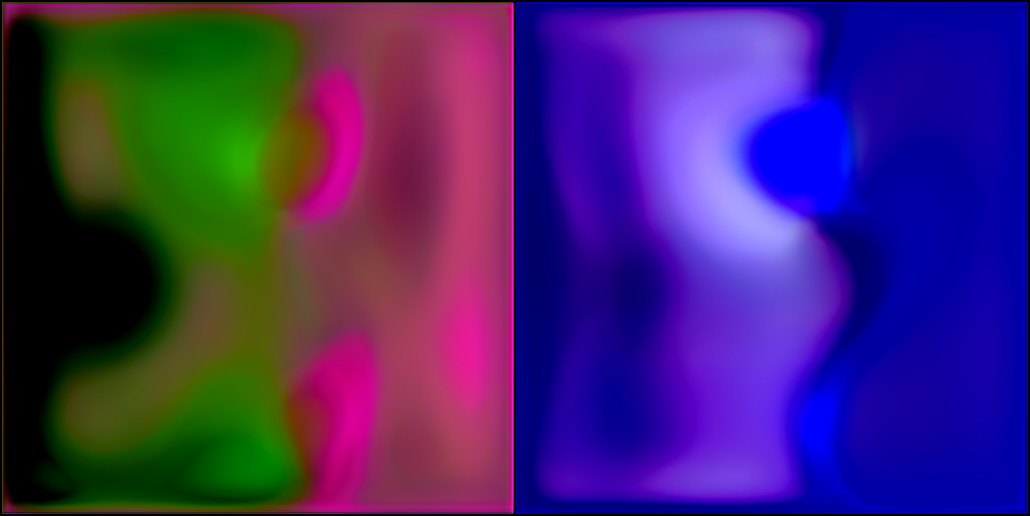}
  \caption{Still from \textit{(un)stable equilibrium 1:4}. The work can be viewed at: \url{https://youtu.be/SL69OZD_-cM}}
  \label{fig4}
\end{figure}

\newpage

\begin{figure}[!htb]
  \centering
\includegraphics[width=0.9\textwidth]{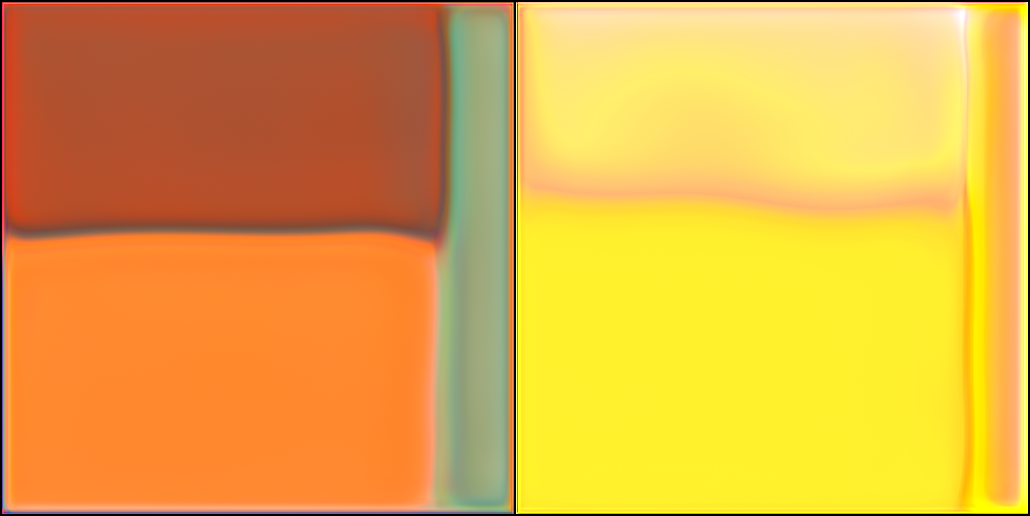}
  \caption{Still from \textit{(un)stable equilibrium 1:5}. The work can be viewed at: \url{https://youtu.be/b8X-KrO4JzM}}
  \label{fig5}
\end{figure}

\begin{figure}[!htb]
  \centering
\includegraphics[width=0.9\textwidth]{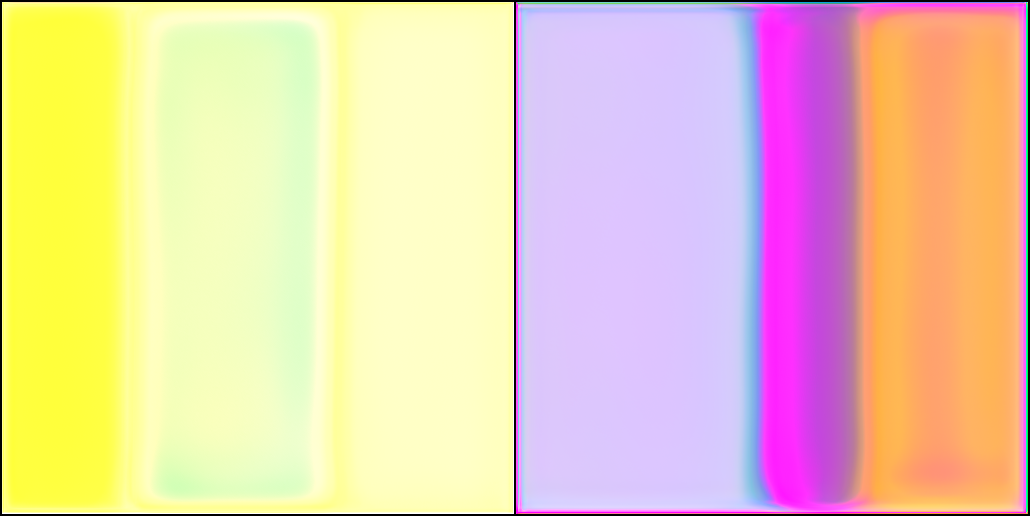}
  \caption{Still from \textit{(un)stable equilibrium 1:6}. The work can be viewed at: \url{https://youtu.be/Jxhi3P2edVQ}}
  \label{fig6}
\end{figure}

\end{document}